\documentclass[11pt]{article}
\usepackage[final]{acl}

\usepackage{amsmath}
\usepackage{arydshln}
\usepackage{graphicx}
\usepackage{subfigure}
\usepackage{tikz}
\usepackage{multicol}
\usepackage{multirow}
\usepackage{xcolor}
\usepackage[normalem]{ulem}
\usepackage[]{algorithm2e}
\usepackage{verbatim}

\usepackage{times}
\usepackage{latexsym}
\usepackage[T1]{fontenc}
\usepackage[utf8]{inputenc}
\usepackage{microtype}

\title{RNN Transducers for Nested Named Entity Recognition with constraints on alignment for long sequences}

\author{
Hagen Soltau, Izhak Shafran, Mingqiu Wang \& Laurent El Shafey\\
Google Brain\\
{\tt soltau,izhak,mingqiuwang,shafey@google.com}\\
\vspace*{0.8cm}}

\date{January 2022}

\begin{document}
\maketitle

\begin{abstract}
Popular solutions to Named Entity Recognition (NER) include conditional random fields, sequence-to-sequence models, or utilizing the question-answering framework. However, they are not suitable for nested and overlapping spans with large ontologies and for predicting the position of the entities. To fill this gap, we introduce a new model for NER task -- an RNN transducer (RNN-T). These models are trained using paired input and output sequences without explicitly specifying the alignment between them, similar to other seq-to-seq models. RNN-T models learn the alignment using a loss function that sums over all alignments. In NER tasks, however, the alignment between words and target labels are available from the human annotations. We propose a fixed alignment RNN-T model that utilizes the given alignment, while preserving the benefits of RNN-Ts such as modeling output dependencies. As a more general case, we also propose a constrained alignment model where users can specify a relaxation of the given input alignment and the model will learn an alignment within the given constraints. In other words, we propose a family of seq-to-seq models which can leverage alignments between input and target sequences when available. Through empirical experiments on a challenging real-world medical NER task with multiple nested ontologies, we demonstrate that our fixed alignment model outperforms the standard RNN-T model, improving F1-score from $0.70$ to $0.74$.  
\end{abstract}

\section{Introduction}
\label{sec:intro}
Named entity recognition (NER), the task that consists of locating and identifying entities such as cities, quantities, or people within a text, plays a critical role in understanding natural language~\cite{manning99foundations}.
It is often one of the first steps within a text processing chain, and reaching a high accuracy is, hence, critical to avoid a negative impact on the downstream tasks. Over the last few decades, there has been a considerable amount of work, both in academia and industry, to build robust NER systems.
However, NER task still faces several challenges, notably in the presence of overlapping entity spans. For instance, entities may be nested as in \emph{University of California, Berkeley}, which is a \emph{university name} with a nested \emph{state} (California) and \emph{city} (Berkeley) and the model is required to predict all three labels.

In this paper, we propose a novel approach for NER that handles nested or overlapping spans. Specifically, the key component of our NER system is an RNN transducer (RNN-T), a model that was originally proposed for speech recognition. While RNN-T models for speech recognition have to learn the alignment between input and target sequences, we propose a key modification for the NER task by fixing the alignment with the given human annotations. This change significantly reduces the hurdle when training over long sequences with limited data. Additionally, we provide a method to relax the alignment between the two extremes of fixed and unconstrained alignment (Section~\ref{sec:rnnt}).
After introducing our experimental setup (Section~\ref{sec:exp_setup}), we demonstrate the benefits of constraining the alignments by empirical experiments on a challenging real-world medical NER task with multiple nested ontologies (Section~\ref{sec:experiments}).
Finally, we show the relationship between this work and a prior seq-to-seq model applied to NER (Section~\ref{sec:seqtoseq}).

\begin{figure}
 \centering 
 \includegraphics[scale=0.4, trim= 0.5in 3.5in 1in 2in]{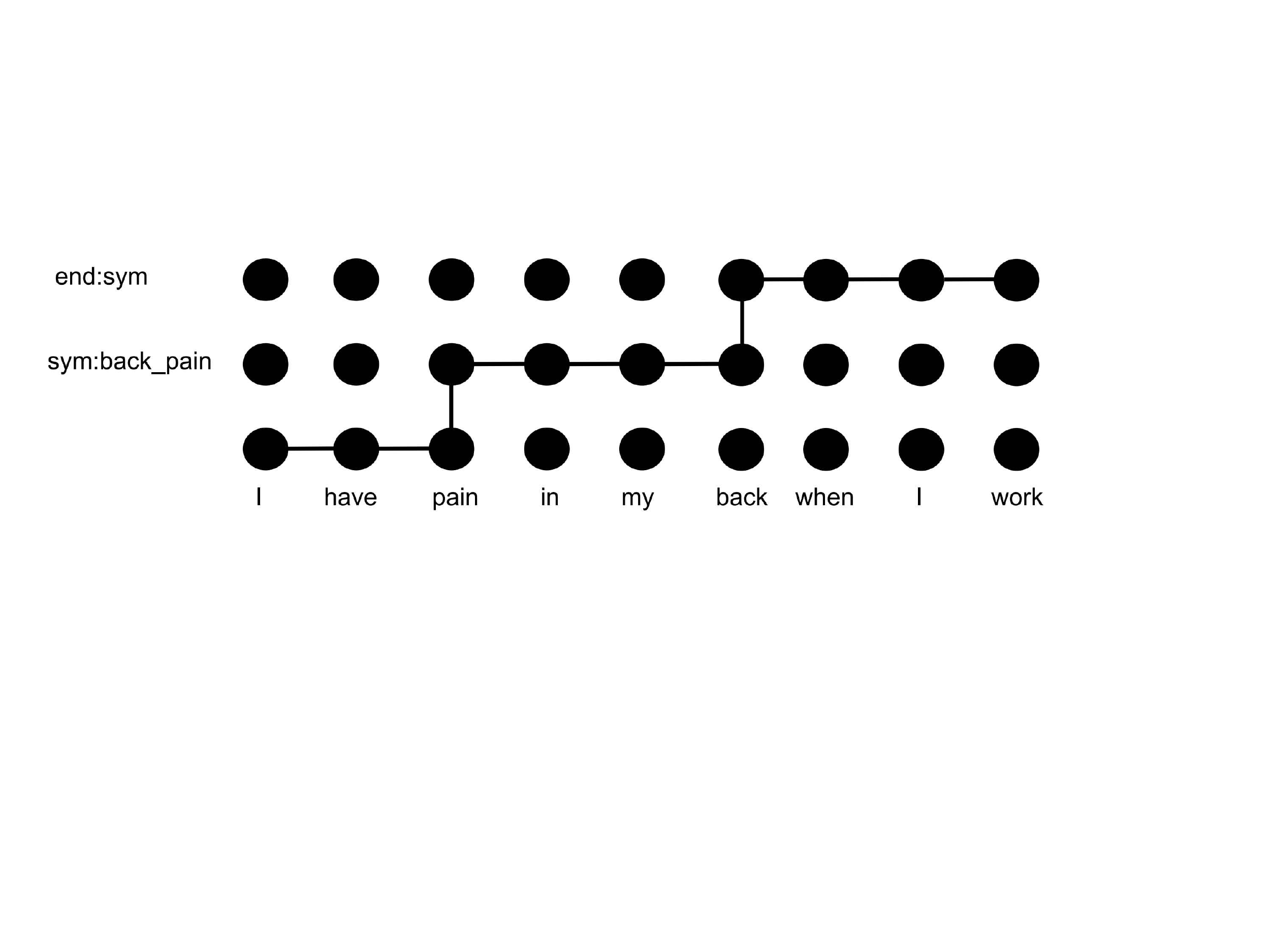}
 \caption{A RNN-T trellis illustrating the alignment of input {\em pain in my back} to the label {\em sym$\colon back\_pain$} where {\em end}-markers indicates the end of a span. Horizontal lines denote time steps in the input sequence (with {\em blank} output symbols) and vertical lines indicate generating the next label in the target sequence.}
 \label{fig:rnnt}
\end{figure}

In summary, the contributions of our paper are:
\begin{itemize}
    \item We introduce RNN transducers for NER tasks and show that they are well suited for predicting position and label for nested and overlapping entities with large ontologies.
    \item We introduce fixed and constrained alignment losses that better utilize human annotations and improve F1 score from 0.70 to 0.74. The fixed alignment loss provides better data efficiency and generalization to long sequences.
    \item We show that typical RNN-T models with unconstrained alignment loss can also benefit from unlabeled data. 
    \item While many public tasks utilize short and well-segmented inputs, we report experiments on a difficult real-world task with both short and long sequences to highlight the challenges associated with long sequences.
\end{itemize}

\section{Related Work}
\label{sec:background}
Linear-chain conditional random fields (CRF) has proved to be a reliable workhorse for NER. In CRFs, each input token is mapped to an output label and the output dependencies are modeled via a transition matrix. However, the CRFs are not designed to generate multiple targets for a given input token and, hence, are not well-suited for tasks with nested spans with multiple labels. 

A simple approach is to modify the label set and use standard CRFs. The new label set is created by taking the cross-product of all possible labels that can be tagged on a single input token. This does not scale well due to the explosion in the label space and the resulting data sparsity. The data sparsity is particularly acute in NER tasks since the typical NER corpora are not very large, resulting in substantial performance degradation (e.g.,~\cite{Nan-SAT,strakova-etal-2019-neural}).

Another approach splits the task as hierarchical inference where spans relevant to all the labels are detected first and then the spans are classified into labels. For example, this hierarchical approach was successfully demonstrated in labeling street scenes~\cite{huang-crf} and in labeling symptoms in medical conversations~\cite{Nan-SAT, du-etal-2019-learning}.

    
The recent interest in large models has motivated researchers to cast translation,  summarization and other NLP problems as question answering (QA) tasks~\cite{mccann2018natural}. Framing NER task as QA task is not straightforward and adds more complexity~\cite{li-etal-2020-unified}. The query prompts are natural language definitions of the labels, as in the annotation guidelines. All the entities in the input for a given label are inferred using a two-stage process -- detecting start/end positions of the spans using binary classifiers for each input position and then classifying the spans into valid/invalid ones. This approach does not scale well at a large label space.   

    
Another approach is to cast NER as a seq-to-seq task, mapping a sequence of words to a sequence of labels (e.g., ~\cite{Kannan2018}). A drawback of this method is that the position of the inferred label with respect to the input sequence is not predicted. This is a consequence of using a soft alignment between encoders and decoders. Without the position information, the model is only solving the NER task partially and is not useful in domains such as medical where explainability is important.
    
A variant of seq-to-seq model alleviates the soft-alignment issue with two modifications~\cite{strakova-etal-2019-neural}. First, the targets are not only the labels, but also include a special {\em eow} (end-of-word) token for every input token. For example, {\em The play is in \underline{New York City}} is represented as {\em The play is in New York City} and {\em eow eow eow eow loc\_beg eow eow eow loc\_end} in the input and target sequences. Second, the model uses {\em eow} tokens to advance the decoder and predict the label for the input word under consideration, thus uses a hard-alignment between encoders and decoders. With these two modifications, the seq-to-seq model with hard-attention outperforms the LSTM-CRF model. 


Taken together, current models are not well-suited for NER tasks with nested spans. One exception is the {seq-to-seq model with hard-attention} and we will compare our model with this work.
\section{RNN Transducer}
\label{sec:rnnt}
For context, we describe RNN Tranducer (RNN-T), a sequence transduction model that was designed for speech recognition tasks~\cite{Graves2012,Graves2013}. Though the model consists of encoder (transcription) and decoder (prediction) networks, it differs from seq-to-seq in how models learn alignments between the input and the output sequences. For RNN-T, this is achieved by relying on a cost function that sums over all possible alignments, computed via forward-backward (or Baum-Welch) algorithm, similar to Hidden Markov Models (HMMs).




\subsection{Model Components} \label{sec:align_prob}
The {\em transcription} (encoder) network maps the input sequence to an abstract representation using a stack of bi-directional layers, which in our case consists of transformers with restricted context~\cite{qian2020, Vaswani2017}. The {\em prediction} (decoder) network is akin to language model in speech recognition and is typically a uni-directional LSTM. One important distinction from seq-to-seq models is that the decoder does not use soft attention over the encoded vector sequence. Instead, the model relies on an augmented {\em blank} label symbol with hard-attention. A {\em joint} network combines the $t$-th vector from the {\em transcription} network and the $u$-th vector from the {\em prediction} network to compute the probability $P(k|t,u)$ of $k$-th label for the $(t,u)$ alignment. The two vectors are combined using a dense layer with $\tanh$ activation and subsequently projected to the output vocabulary.  

\subsection{Loss Computation Over An Alignment}
When the model emits the {\em blank} symbol, the input position is advanced from the current position $t$ to the next in alignment, as illustrated by the trellis in Figure~\ref{fig:rnnt}.  Thus, a valid path in the alignment trellis is of length $T+U$ where $T$ and $U$ are the lengths of the inputs and output sequences. The model can emit multiple output labels at any given input position. The output labels from the model except for the {\em blank} symbol are fed back as input to the {\em prediction} network. Note, this is an important distinction from the seq-to-seq model with hard-attention mentioned in Section~\ref{sec:background}.

\subsection{Model Training and Inference}
During training, the loss of a regular RNN-T model is computed by marginalizing over all valid alignments between input $x$ and output $y$ sequences.\\[-2mm]
\begin{equation}
   L_{\operatorname{regular}} = P(y \mid x) = \sum_{A} P(A \mid x)
\end{equation}\\[-6mm]

The total probability for a given alignment $P(A)$ can be calculated by summing up the probabilities $P(k \mid t,u)$ for every step in the path. The total probability $P(y \mid x)$ can be computed efficiently by forward-backward algorithm on accelerators as matrix products~\cite{Sim2017, Bagby2018}.

During inference, the {\em frame synchronous} beam search technique, outlined in Algorithm~\ref{fig:rnnt_search}, allows for efficient pruning of the search space. This algorithm is more efficient than the original RNN-T search~\cite{Graves2012} and clearly illustrates how the predicted labels align with the input sequence, making it suitable for named entity prediction. 
\DecMargin{1em}
\begin{algorithm}[t]
 paths = []\\
 \For{t in 1..T} {
    new\_paths = []\\
    \For{u in 1..max\_expansion}{
        \For{path in paths} {
            remove path from paths\\
            \For{label in vocab} {
                ext\_path = path + [label]\\
                prob = CalcProb(path, label, t)\\
                \If{prob $\geq$ best - beam} {
                    \eIf{label == blank} {
                        new\_paths.append(path)
                    } {
                        paths.append(ext\_path)
                    }
                }
            }
        }
    }
    paths = new\_paths
 }
\vspace{0.1in}
 \caption{Frame-synchronous beam search for RNN-T. Details like maintaining the best score and search space pruning are omitted. The key is that the algorithm works in a breath-first manner and keeps a stack of {\em comparable} paths for efficient pruning.}
 \label{fig:rnnt_search}
\end{algorithm}
\IncMargin{1em}

\subsection{Fixed Alignment Loss}
As in other seq-to-seq models, RNN-T models are typically trained using paired input and output sequences without explicitly specifying the alignment between them. The model learns the alignment using a loss function that sums over all alignments, which makes them attractive for ASR where word-level alignments are unavailable.

In NER tasks, however, the alignment between words and target labels are available from the human annotations. There is no need for the model to learn the alignment and we can simplify the task by modifying the loss to a {\em fixed-alignment loss}. \\[-2mm]
\begin{equation}
   L_{\operatorname{fixed}} = P(y \mid x) = P(A \mid x)
\end{equation} \\[-6mm]
We explicitly encode the alignment information in the input and compute the loss with respect to the {\em given} alignment only. This is in particular useful when training data is sparse as it is often the case for NER tasks. As we demonstrate in experiments, this improves learning when training with long sequences, which in turn allows us to increase the contextual information available to the model and improve accuracy on the NER tasks.

\subsection{Constrained Alignment Loss}
Filling the gap between the two extremes of the original {\em unconstrained alignment} loss and the new {\em fixed-alignment} loss, we propose a {\em constrained alignment} loss. The key idea is to allow a user-defined relaxation of the given training alignment. This could be very useful when human annotations are noisy and model can learn perturbations of the given alignment with better likelihood.

We accomplish this by manipulating the label ($y$) and {\em blank} ($b$) matrices of RNN-T models which are employed in the forward-backward algorithm to compute the sum over alignments~\cite{Sim2017, Bagby2018}. Specifically, they correspond to labels ($\log \, P(y_{u}\mid x_{t})$) and blanks ($\log \, P(b_{u}\mid x_{t})$) defined over input and output time steps. Intuitively, the $(t,u)$ entries in the two matrices represent the associated with the horizontal and vertical transitions in the the alignment trellis in Figure~\ref{fig:rnnt}. The {\em fixed-alignment} loss can be viewed as an instance where all the entries of the matrices are set to negative infinity except for those corresponding to the given alignment. 

The {\em constrained alignment} loss is parameterized by user-specified relaxations in $(t,u)$ dimensions of the alignment, from which we create a 2D unit convolution filter ($r$).\\[-2mm]
\begin{equation}
    L_{\operatorname{constr.}} = P(y \mid x) = \sum_{A \in C(delta)} P(A \mid x)
\end{equation}\\[-8mm]

The given training alignment is represented as two Boolean matrices (masks) corresponding to the alignment in the $y$ and $b$ matrices. The masks are then expanded (relaxed) to a set of valid alignments $C(delta)$ using a 2D convolution with a unit rectangular filter, whose dimensions are $1 + 2*delta(t)$ and $1 + 2*delta(u)$. Thus, the degree of relaxation from fixed alignment is controlled by $delta(t)$ and $delta(u)$ and we use the same $delta$ for both dimensions in our experiments. The expanded Boolean masks are then intersected with the $y$ and $b$ matrices and all the entries not masked are set to negative infinity. Once the alignment constraints are incorporated into the $y$ and $b$ matrices, we compute the forward-backward algorithm to obtain the {\em constrained alignment} loss.

\section{Experimental Setup}
\label{sec:exp_setup}
We report experimental results on a real-world NER task on medical conversations ~\cite{shafran-etal-2020-medical} that contains overlapping and nested entities with multiple ontolgies and large label sets. 

\subsection{Corpus}
The corpus contains about 100K unlabelled conversations between medical providers and patients. Of these about 6k conversations were labeled by professional scribes and correspond to 5 ontologies -- medications, symptoms, conditions, diagnoses and treatments. In all, the corpus contains about 100k medications, 64k symptoms, 42k conditions, 38k diagnoses and 6k treatments. The corpus was split into about 5k conversations for training and 500 conversations for development. For more details, see~\cite{shafran-etal-2020-medical}.


\begin{table}[h]
\centering
\begin{tabular}{|l|c|} \hline
Ontology & training examples  \\ \hline
Medications & 100,000 \\ \hline
Symptoms & 64,000 \\ \hline
Conditions & 42,000 \\ \hline
Diagnoses & 38,000 \\ \hline
Treatments & 6,000 \\ \hline
\end{tabular}
\caption{Corpus Ontologies}
\label{tab:data_stats}
\end{table}

\begin{figure}
 \centering 
 \subfigure{\includegraphics[width=3.7cm, trim= 0.5in 0.0in 0.0in 0.0in]{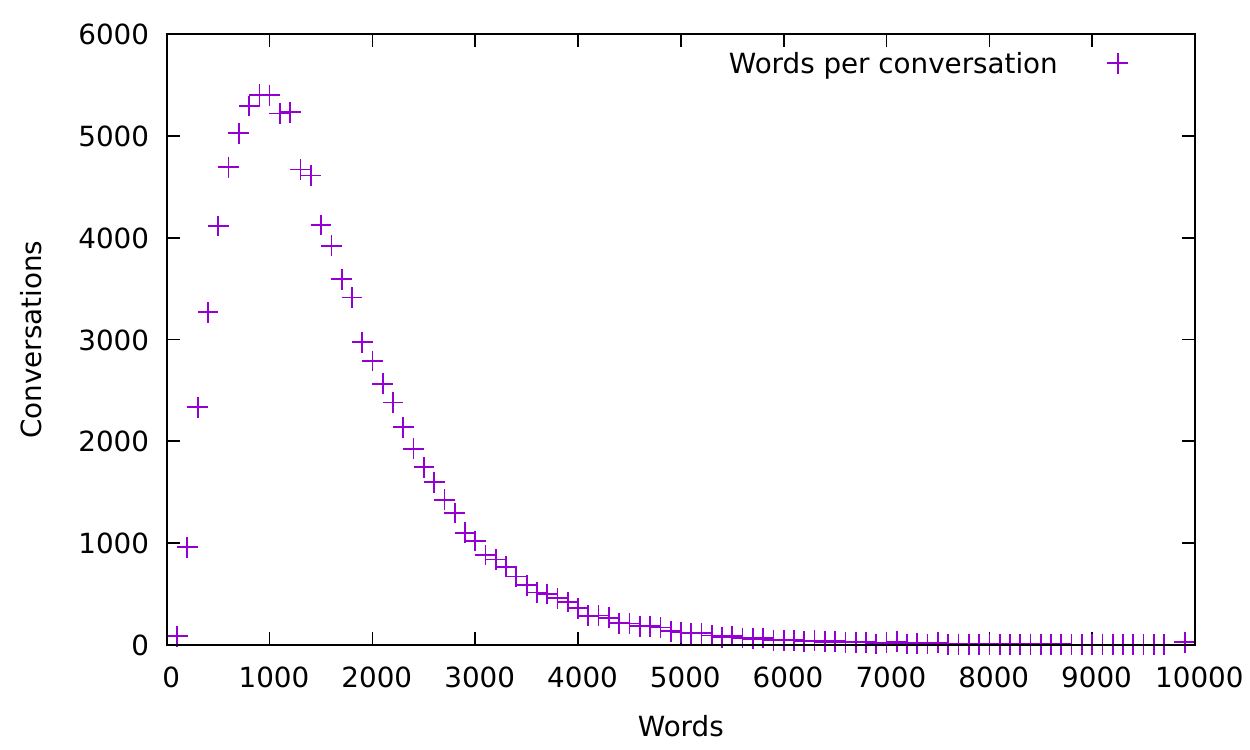}}
\subfigure{\includegraphics[width=3.7cm, trim= 0.0in 0.0in 0.5in 0.0in]{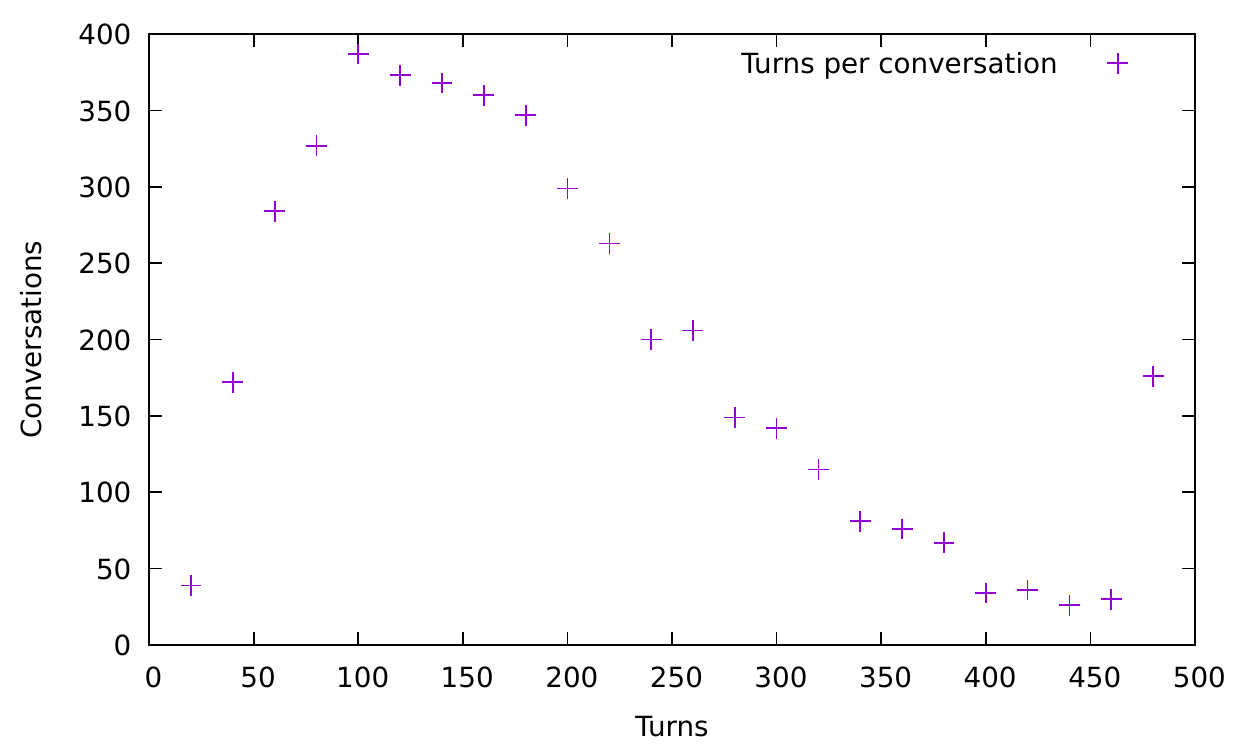}}
 \caption{Corpus statistics. Distribution of number of
 words (left) and speaker turns (right) per conversation. Sentences have an average length of 10 words and there are about 160 speaker turns per conversation and each conversation has on average 1600 words.}
 \label{fig:data_stats}
\end{figure}

Figure~\ref{fig:data_stats} contains data statistics that highlight the conversational nature of the data. On average, there are 160 speaker turns and certain conversations have up to 500 turns. While many sentences are very short, there is a long tail of sentences with 50 or more words. Overall, the high variability in number of turns and sentence lengths makes it very difficult to segment conversations into smaller chunks without loosing context. The need to accommodate long sequences in training and inference makes this a challenging task.



\subsection{Evaluation Metrics}
A common evaluation metric for named entity recognition (NER) model is {\em F1 score}. Traditionally, the {\em F1 score} measures the correctness of the predicted labels and the location of input text over which the labels are predicted.  More recently and often in the context of seq-to-seq models, NER models are evaluated using {\em global F1 score} which only evaluates the predicted labels, without considering their locations. This is convenient for seq-to-seq models since their label sequences are not aligned with the input sequence. However, the {\em global F1 score} does not accurately reflect the performance of the NER model in practical applications where location is important. For example, in medical applications, location is important for {\em explainability} of the model output for the end users. Therefore, we report the traditional or {\em local F1 score} that measures the correctness of both the predicted labels and their locations.

\subsection{Pre-Processing}
The average sequence length of our conversations is about $1600$ words, with a long tail of very long conversations as it can be seen in Figure~\ref{fig:data_stats}. While it is quite common to break conversations by sentence end markers, this approach does not work well in our use case for the following reasons. First, the model is intended to be used with audio as input modality and the current ASR models do not predict sentence end markers sufficiently well~\cite{Soltau2021}. Second, our data is conversational in nature, and many sentences are very short (e.g. answering questions, yes/no, etc). A single sentence will often not provide sufficient context to predict medical entities and choosing a fixed window of sentences might break the context at the wrong boundaries.

Instead, we feed entire conversations to our model at inference time. This avoids breaking context at fixed intervals. However, given the long tail of long conversations and limited memory on accelerator devices, during training time we have to break conversations. As we do not have good break points, we randomly cut segments from conversations. The segment lengths of the training examples are between $280$ to $320$ words. This randomized cutting is performed when each example is loaded during training. As the training progresses, the model essentially sees different views of the data; this can also be seen as a form of data augmentation. During the test time, the full conversations are fed as input sequence.

For tokenization, we use Morfessor~\cite{Virpioja2013} to split words into morpheme like units and use a {\em space} marker to encode word boundaries. The entire vocabulary consists of $15551$ tokens, many of them forming entire words. On average, a word is split into $2.2$ tokens.

\subsection{Pre-Training}
The text encoder is a Transformer-XL~\cite{dai2019transformerxl} architecture with 15 layers and has in total 473m parameters. Each layer uses an attention context of $\pm 20$ tokens and the entire context of the encoder is then 600 tokens.

The pre-training corpus consists of $100k$ unlabeled medical conversations with $177m$ words. We apply the same pre-processing for pre-training and training and randomly cut segments of $280$ to $320$ words.

For pre-training of the encoder, we mask input tokens similar to the approach in BERT~\cite{devlin2019bert}, but the model needs to learn not only to recover the masked inputs, but also the entire sequence including the unmasked tokens. This is done to encourage the encoder to learn words as well as context representations. The masking operation is done at word level before tokenization, so that entire words are masked out. This is to avoid cases where non masked out word fragments would make it easy to recover words. The model needs to learn to use context for a given mask token to generate all word pieces that belong to that masked out word. This means also the input and target sequences are not of the same length, and we use a RNN-T model to train on this data. We use a single LSTM layer for the prediction network of the RNN-T as only the pre-trained encoder is then used for the down-stream tasks.

\section{Experiments}
\label{sec:experiments}

\subsection{Unconstrained Alignment Loss}

\begin{figure}
\subfigure[Unconstrained Align.]
{\includegraphics[scale=0.29, trim= 0.0in 0.0in 0.0in 0.0in]{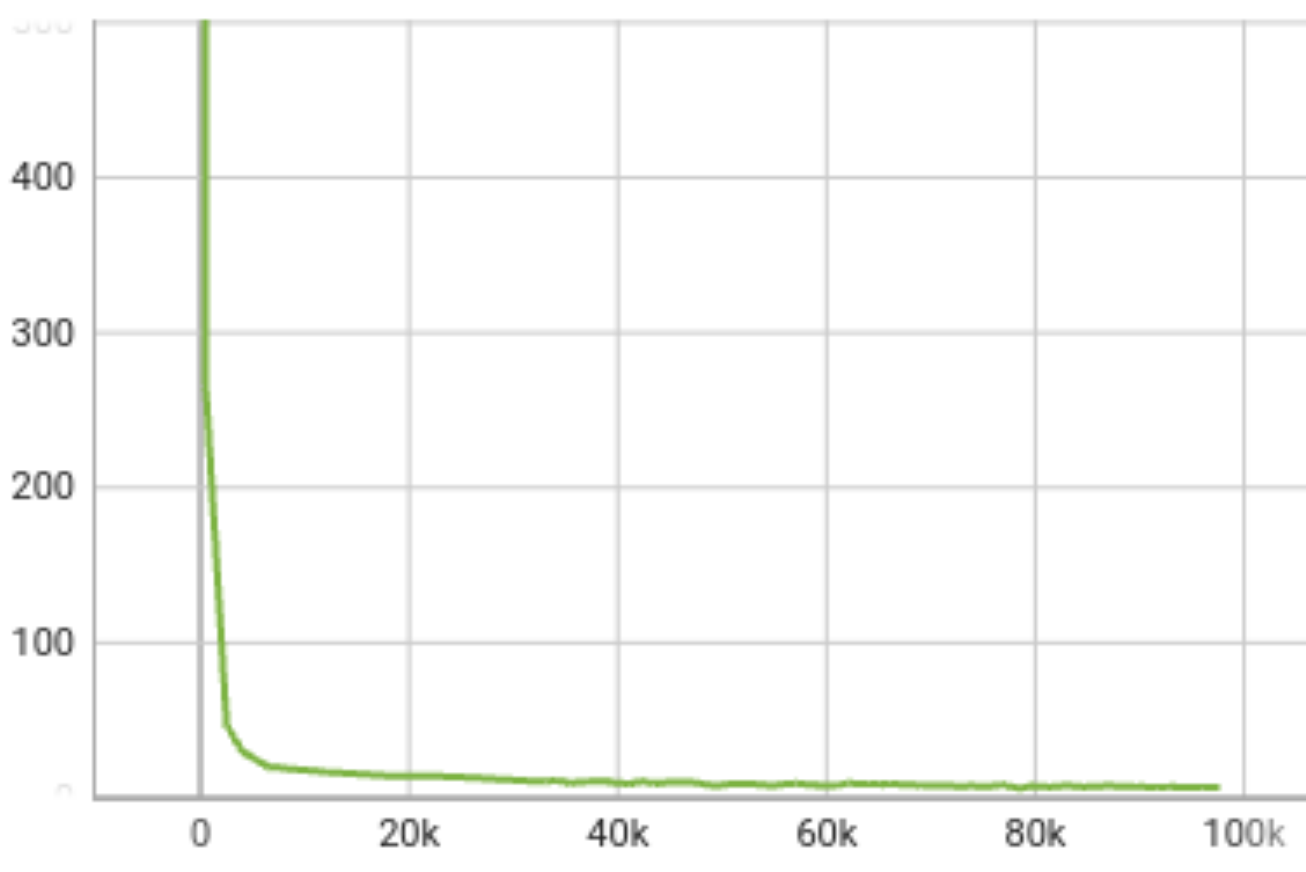}}
\subfigure[Fixed Align.]
{\includegraphics[scale=0.29, trim= 0.0in 0.0in 0.0in 0.0in]{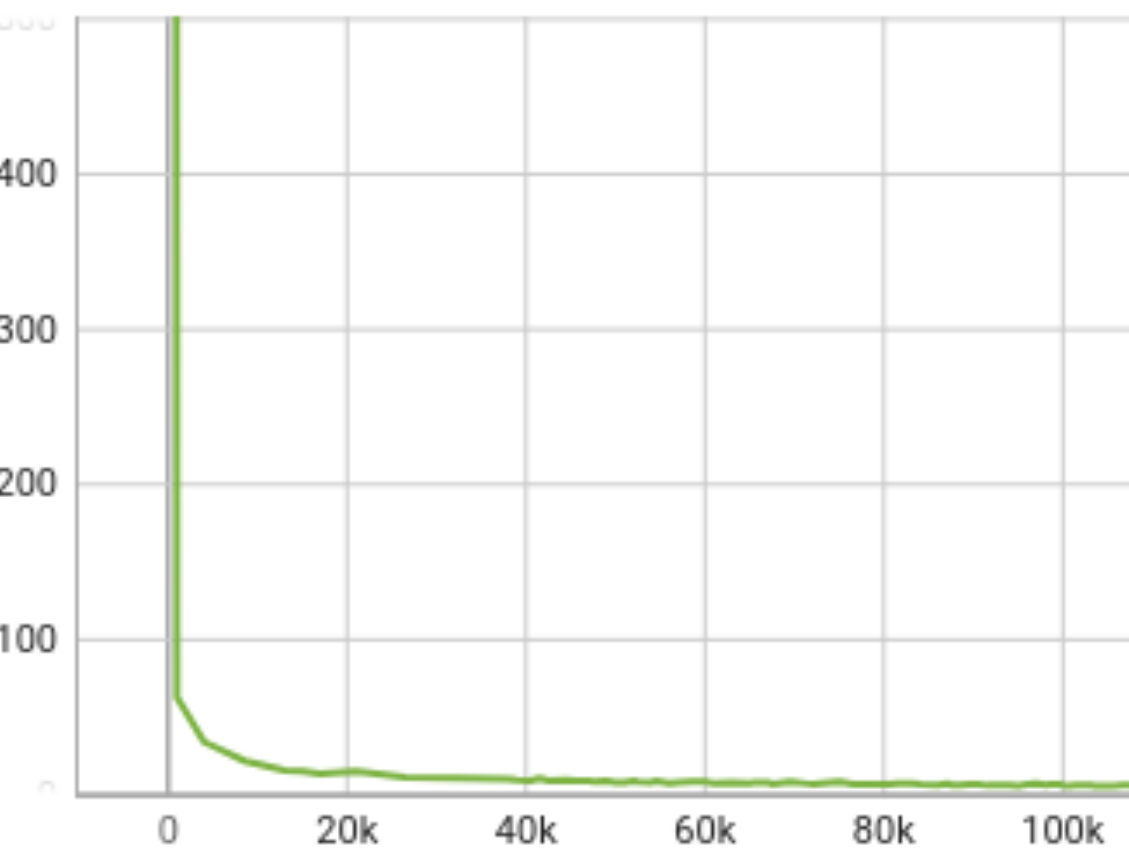}}
\caption{Negative Log Likelihood. Training loss converges very similarly for both unconstrained and fixed alignment models. This is in contrast to the convergence in terms of F1 metrics (figure~\ref{fig:f1}). With unconstrained alignments, the model does not learn to predict entities well while still minimizing the loss function.}
\label{fig:nll}
\end{figure}

The results with a standard RNN-T and an unconstrained alignment using Forward-Backward training are shown in Figures~\ref{fig:nll} (Likelihood) and~\ref{fig:f1} (F1 score). The plot shows that the training works as intended, minimizing negative log-Likelihood loss on the training data. The left plot in 
Figure~\ref{fig:f1} shows the performance at recognizing named entities, both for training as well as test data. While the log-Likelihood of the training data is clearly getting optimized, the model has difficulty learning the task and the learning is unstable as evidenced by the F1 scores on the training data. We have observed that the posterior distribution becomes sharper and sharper during training, minimizing the negative likelihood, while not learning to predict the class labels.

\subsection{Fixed Alignment Loss}
\begin{figure}
\subfigure[Unconstrained Align.]
{\includegraphics[scale=0.29, trim= 0.2in 0.0in 0.0in 0.0in]{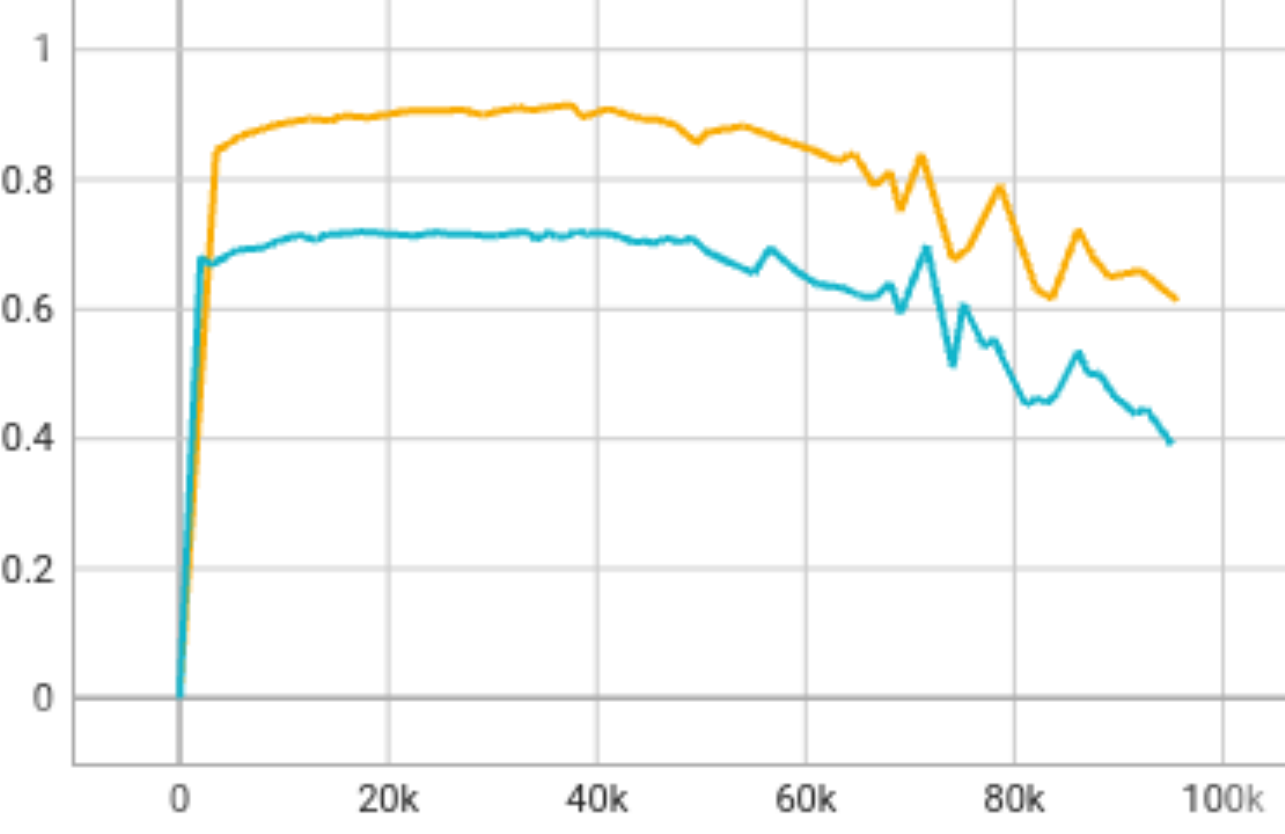}}
\subfigure[Fixed Align.]
{\includegraphics[scale=0.29, trim= 0.0in 0.0in 0.2in 0.0in]{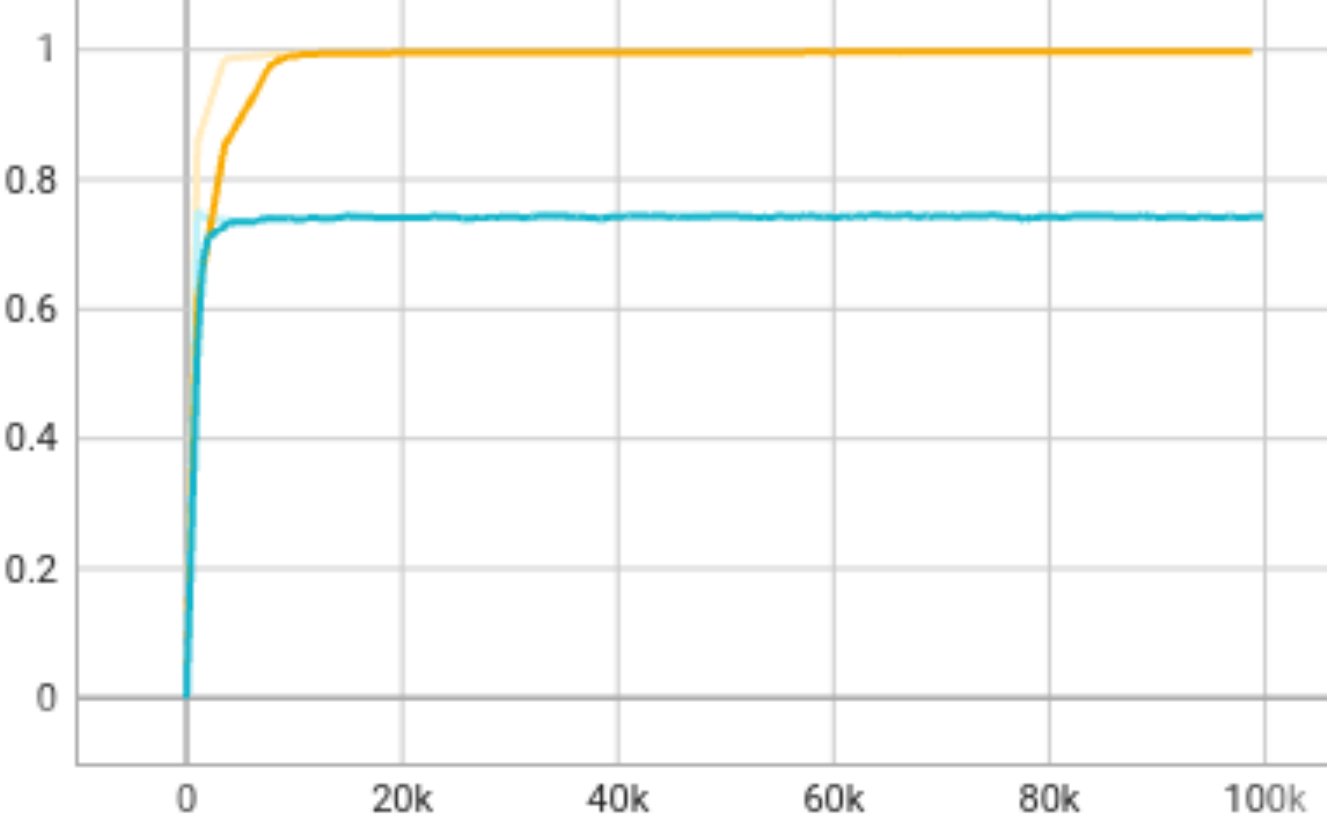}}
\caption{F1 metrics for train (orange) and test (blue) data. Training with unconstrained alignment leads to unstable NER performance, even though Log Likelihood converges well in training.}
\label{fig:f1}
\end{figure}

The advantages of fixed alignment training can be seen in Figure~\ref{fig:f1} comparing side-by-side fixed and unconstrained alignment models, both of which used the same inference setup with regular RNN-T beam search. While the log-likelihood loss on training data converges similarly as in unconstrained alignment loss, the performance in terms of F1 scores tracks nicely with the likelihood and the model learns to predict labels. When the loss is computed over fixed alignments, the task of label prediction becomes much easier.\\
The overall F1 scores improve from 0.702 (Precision=0.705, Recall=0.66) with unconstrained alignment loss to 0.743 (Precision=0.789, Recall=0.702) with fixed alignment loss. 

\subsection{Constrained Alignment Loss}
\begin{figure}
 \centering 
 \includegraphics[width=4.8cm]{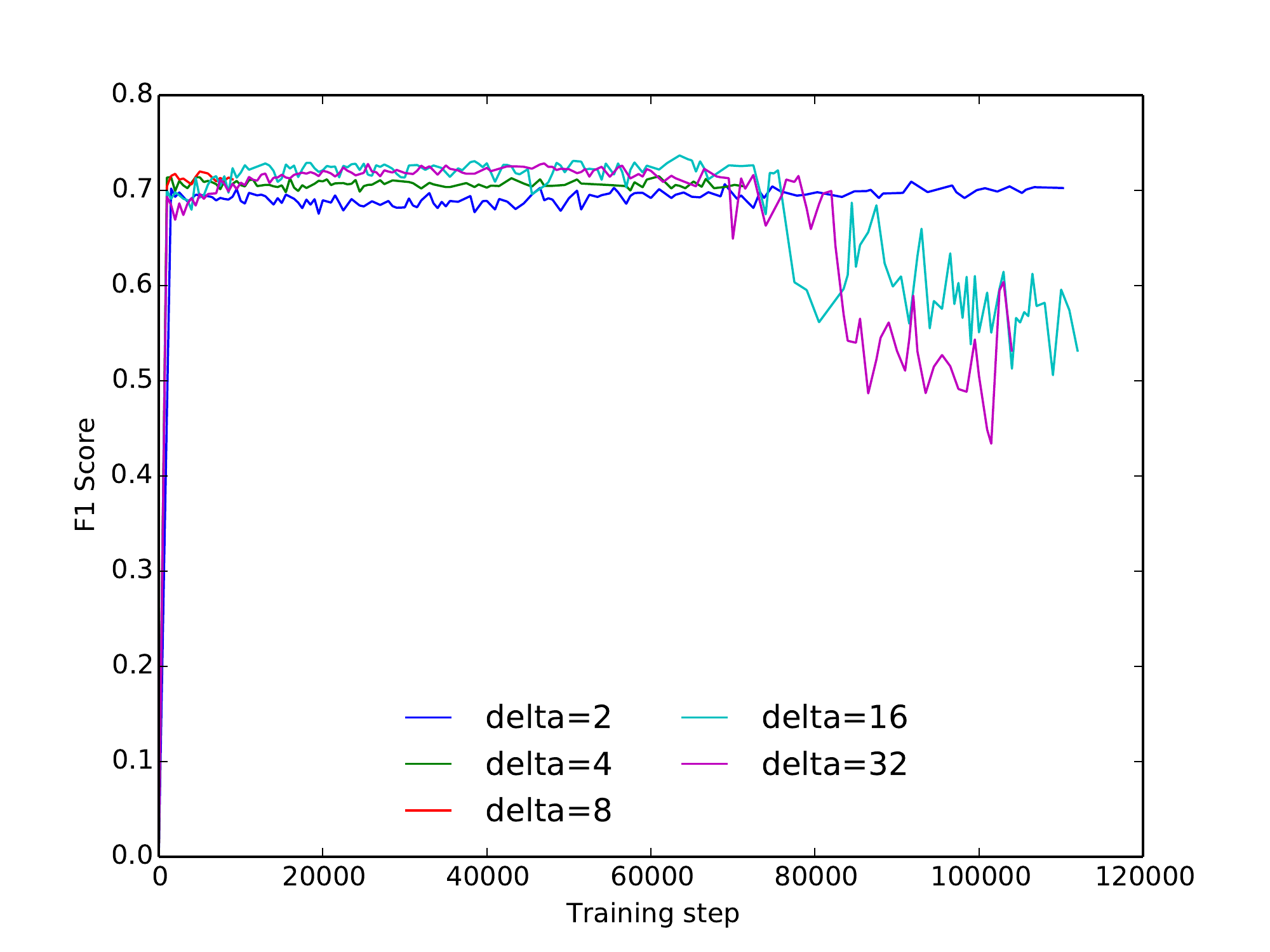}
 \caption{Constrained alignment loss: when the constraints are relaxed, the performance ranges from that of fixed alignment loss to unconstrained alignment loss.}
 \label{fig:constrained}
\end{figure}

For studying the impact of relaxing the constraints on the {\em given} training alignment, we trained models with different alignment tolerances in $(t,u)$ axes. We specified equal tolerance on both axes with values of 2, 4, 8, 16 and 32. As evident from the Figure~\ref{fig:constrained}, when the permitted alignments are close to the {\em given} training alignment, the model performs similar to fixed alignment loss. When the alignments are allowed to diverge from the {\em given} training alignment, the performance resembles that of unconstrained alignment. This demonstrates how constraints in the alignment loss can be effectively used to steer the behavior of the learned model. 

\subsection{Semi-Supervised Learning}
Named Entity recognition models are often trained with limited amounts of labeled data since manually labeling spans of input text is labor intensive. In our case, only 5k of the 100k conversations are labeled. 
In such settings, semi-supervised learning often provides additional gains when the initial model can produce reasonably accurate labels. For evaluating gains from semi-supervised learning, we generate pseudo-labels on the remaining 95k unlabeled conversations and fold them back into the training data.

\begin{table}[h]
\centering
\begin{tabular}{|l|c|c|} \hline
Loss & 5k labeled & + 100k unlabeled \\ \hline
Unconstrained & 0.702 & 0.747 \\ \hline
Fixed         & 0.743 & 0.762  \\ \hline
\end{tabular}
\caption{Adding unsupervised training data gives substantial gains for the unconstrained alignment case, where the added training data helps the model to learn aligning words to labels.}
\label{tab:unsupervised1}
\end{table}

\begin{figure}
 \centering 
 \subfigure[Unconstrained Align.]{\includegraphics[width=3.7cm, trim= 0.5in 0.0in 0.0in 0.0in]{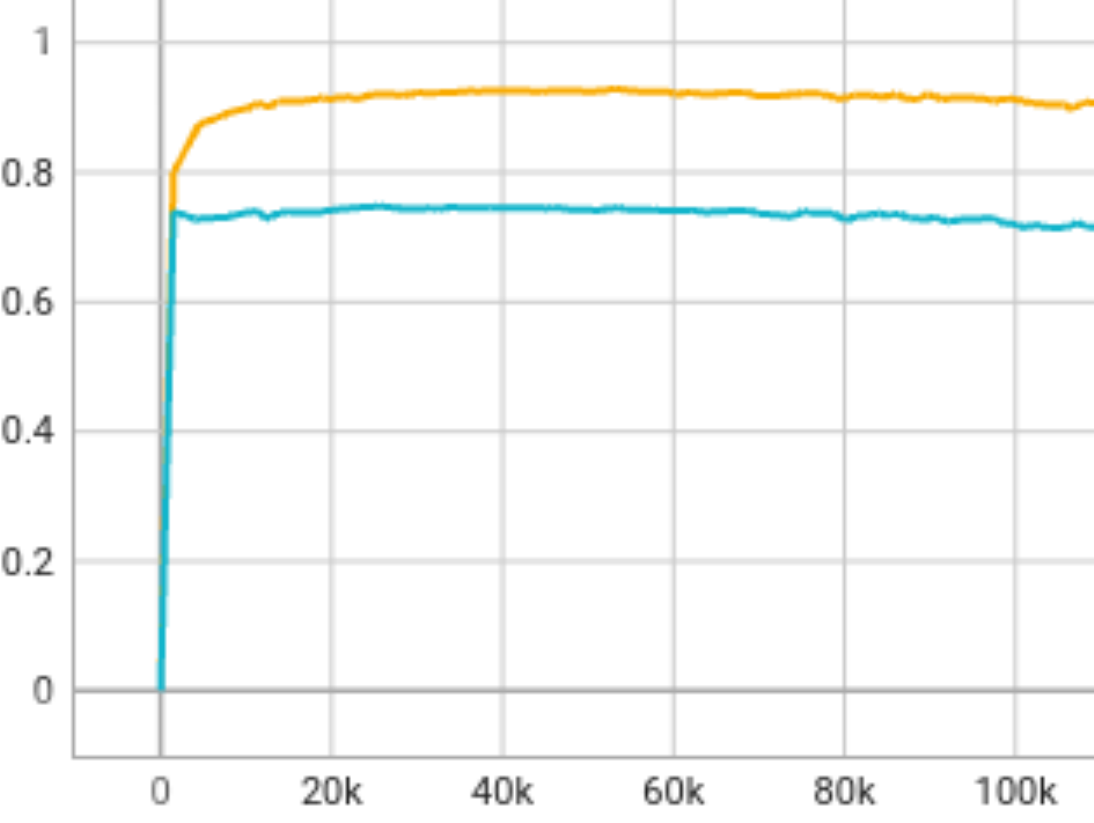}}
 \subfigure[Fixed Alignment Loss]{\includegraphics[width=3.7cm, trim= 0.0in 0.0in 0.5in 0.0in]{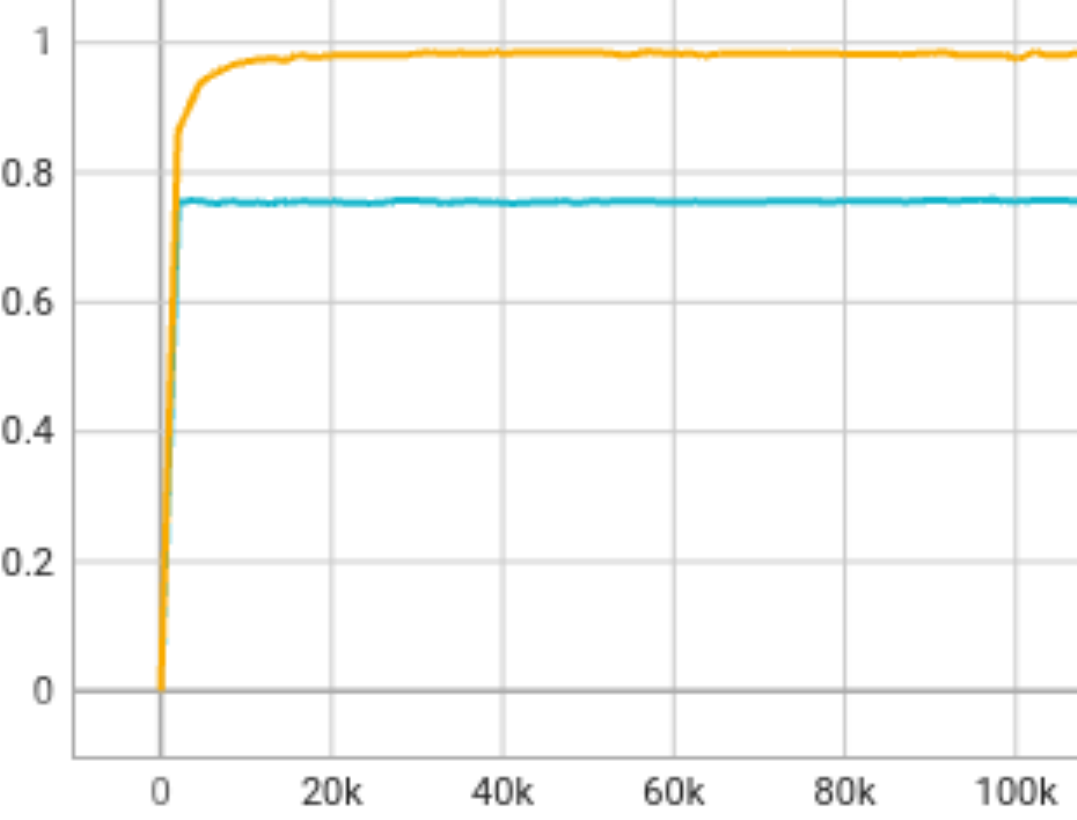}}
 \caption{Setup with additional unsupervised data.
 F1 scores for training (orange) and test (blue) data. The plot shows training convergence of unconstrained and fixed alignment losses when a large amount of additional unsupervised data is used. The additional data makes unconstrained loss much more robust, see Figure~\ref{fig:f1} for comparison when no unsupervised data is used.\\[-6mm]}
 \label{fig:unsupervised}
\end{figure}

Table~\ref{tab:unsupervised1} summarizes the results from adding semi-supervised training data. The additional data benefits models with unconstrained alignment loss more than the fixed alignment loss. If enough data is available,  models can learn alignments without specific human annotations, potentially saving annotation costs. When the data is limited and the model has difficulty learning the alignment all by itself, utilizing the given alignment through fixed-alignment loss is very beneficial. The fixed-alignment loss is so effective that the additional data only brings marginal benefits and the benefits are observed largely for ontologies that have limited training examples, as reported in Table~\ref{tab:unsupervised2}.



\begin{table}[h]
\centering
\begin{tabular}{|l|c|c|} \hline
& 5k labeled & + 100k unlabeled \\ \hline
Overall       & 0.743  & 0.762 \\ \hline
Symptom       & 0.744  & 0.757 \\ \hline
Medication    & 0.894  & 0.904 \\ \hline
Med. Property & 0.643  & {\bf 0.693} \\ \hline
Diagnosis     & 0.722  & {\bf 0.769} \\ \hline
Condition     & 0.732  & 0.747 \\ \hline
Treatment     & 0.586  & 0.607 \\ \hline
Attributes    & 0.611  & {\bf 0.674} \\ \hline
\end{tabular}
\caption{Improvements from semi-supervised training for fixed alignment loss, break-down per ontology. Gains are small for ontologies like Symptoms or Medications that already have many training examples, while larger gains are observed for ontologies like Diagnosis, medical properties and general attributes that had fewer supervised labels.\\[-8mm]}
\label{tab:unsupervised2}
\end{table}

\subsection{Training on Short vs Long Sequences}
\begin{figure}
 \centering 
\subfigure[Segmented test set]{\includegraphics[width=3.6cm, trim= 0.5in 0.0in 0.0in 0.0in]{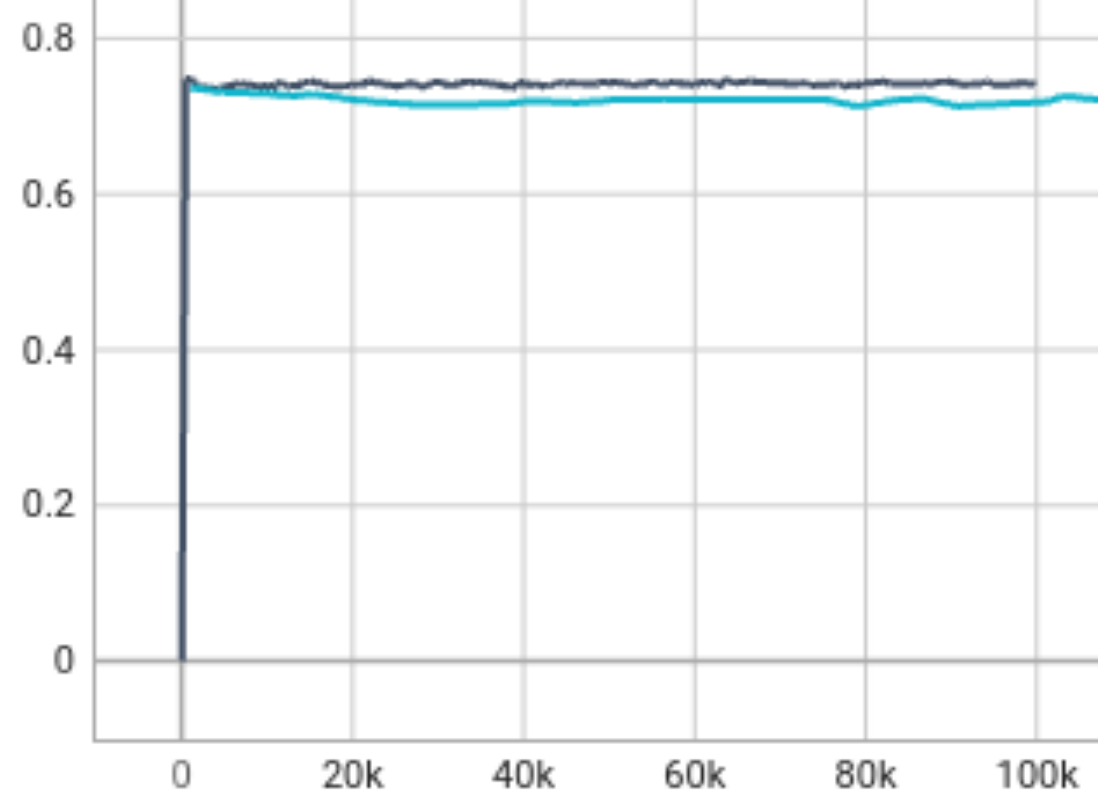}}
 \subfigure[Unsegmented test set]{\includegraphics[width=3.6cm, trim= 0.0in 0.0in 0.5in 0.0in]{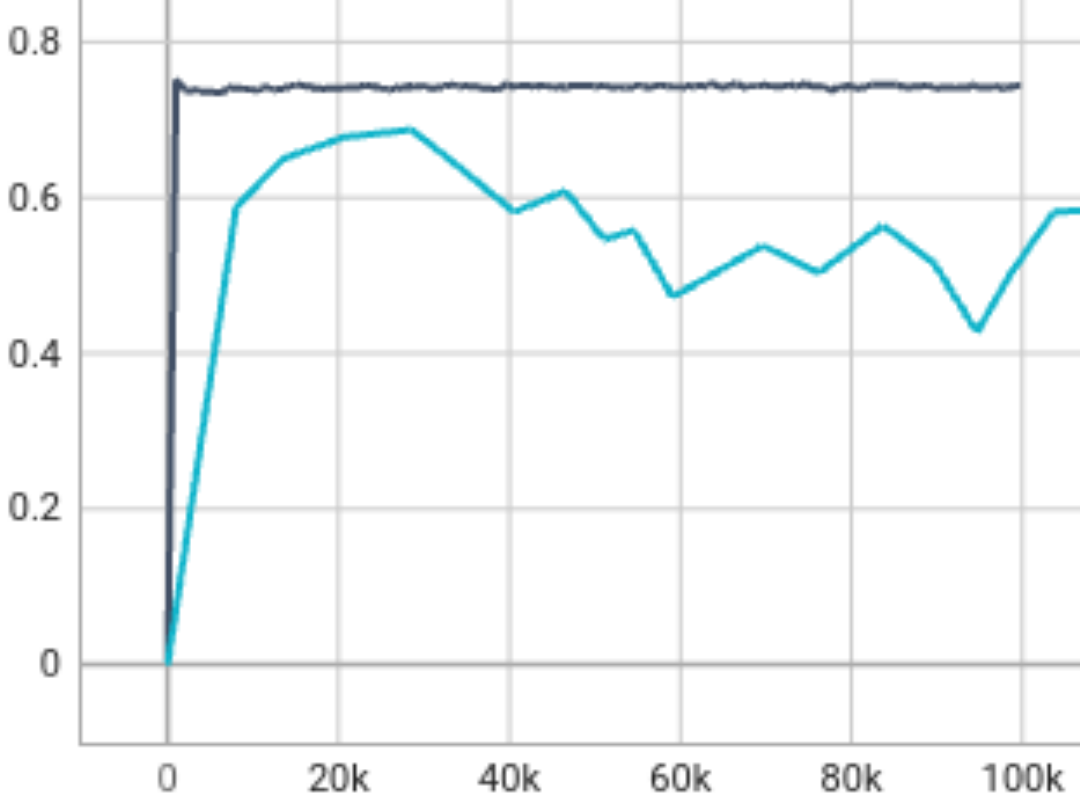}}
 \caption{Fixed alignment model trained on short (50 words, light blue) vs long segments (300 words, dark blue). While the model trained on short segments performs well on {\em segmented} test data, it fails to generalize to a more realistic setting with unsegmented tests data (right plot, light blue curve). In contrast, the model trained on longer sequences generalizes well.\\[-6mm]}
 \label{fig:short_vs_long}
\end{figure}

The difficulty of learning alignments increases with the length of sequences. This is true for modern sequence transduction models (RNN-T, seq2seq), as well as traditional HMM models. In our experimental setting, the raw data are unsegmented conversations and we randomly cut training examples from these conversations during training infeed. To alleviate the training issues with Forward/Backward training (as shown in Figure~\ref{fig:f1}), we could reduce the segment lengths during model training.\\
However, during inference the test data is unsegmented and long. While training on short sequences helps with learning alignments, the models are unable to generalize to long sequences. We demonstrate this by training two models, with 40-60 and 280-320 word segments respectively, both with fixed-alignment loss. Their performance, as shown in Figure~\ref{fig:short_vs_long}, clearly demonstrates poor generalization of the model trained with shorter segments. 

To verify that this is not a training issue and rather a mismatch between train and test conditions, we generated a {\em segmented} version of the test data by applying the same random segment cutting procedure that we apply in training. Each conversation gets segmented multiple times and we average the results of the F1 scores. While this is clearly not a suitable approach when deploying models, it helps us to verify that is indeed a generalization issue where models do not learn to generalize to longer sequences. The results on the segmented test data can be seen in the left plot in Figure~\ref{fig:short_vs_long}. Both models (train on short and long sequences) perform well on segmented data, while only the model trained on long sequences performs well on unsegmented conversations.

\section{Relationship to Seq-to-Seq Models}
\label{sec:seqtoseq}

In section~\ref{sec:background}, we pointed out that general sequence-to-sequence models don't identify the positions of the labels in the input and this solves the NER task only partially.
This limitation was overcome by an approach where the input tokens were tracked using {\em EndOfWord(eow)}~\cite{strakova-etal-2019-neural}, which we will refer to as seq-to-seq with hard-attention.

The approach with hard-attention can be viewed as a special case of RNN-T with fixed-alignment loss. Both models do not use soft-attention that is widely employed in other sequence-to-sequence models. The {\em eow} symbols serve the same purpose as {\em blank} symbols in the RNN-T output sequence where their emission corresponds to advancing to latent representation of the next symbol in the input sequence or the horizontal step in the RNN-T alignment trellis. While the {\em blank} symbol is implicitly encoded in the loss function and not represented in the target sequence, {\em eow} is explicitly represented in the target sequence.  The {\em blank} symbol is not fed back to the decoder as history of model outputs from previous steps, while the {\em eow} is.

While beam search for conventional seq-to-seq model is label synchronous and essentially unrolls a language model,  the encoding of {\em eow} symbols in the output sequence enables the use of frame-synchronous RNN-T beam search and the algorithm shown in Figure~\ref{fig:rnnt_search} can be directly applied.

Aside from the differences in decoder history, there is a small difference in the computation of logits in the two models. In RNN-T, the logits are calculated as follows, whereby $f(t)$ and $g(u)$ represent the outputs from the transcription (encoder) and prediction (decoder) models. \\[-2mm]
\begin{equation*}
\begin{aligned}
    g(u) & = \operatorname{lstm}(u, \operatorname{state})\\
    \operatorname{joint}(t,u) & = W * \tanh(f(t) + g(u)) + b\\
    P(t,u) & = \operatorname{softmax}(\operatorname{joint}(t,u))\\
    \operatorname{state} & = \operatorname{update}(\operatorname{state}, u) \text{ if u} \neq \operatorname{blank}
\end{aligned}        
\end{equation*}\\[-2mm]
In contrast, for a Seq-to-Seq model with hard attention that moves in lockstep with the encoder, the logits are calculated as follows:\\[-2mm]
\begin{equation*}
\begin{aligned}
    \operatorname{attn}(t,u) & = \operatorname{concat}(\operatorname{emb}[u], \operatorname{enc}[t]) \\
    \operatorname{joint}(t,u) & = \operatorname{lstm}(\operatorname{attn}(t,u), \operatorname{state}) \\
    P(t,u) & = \operatorname{softmax}(\operatorname{joint}(t,u)) \\
    \operatorname{state} & = \operatorname{update}(\operatorname{state}, u)
\end{aligned}        
\end{equation*}\\[-2mm]
The main difference is that the combination of encoder and decoder happens before the lstm, while in RNN-T the recurrence occurs only in the language model. The impact of this difference may depend on the task.

\begin{figure}
 \centering 
 \includegraphics[width=3cm]{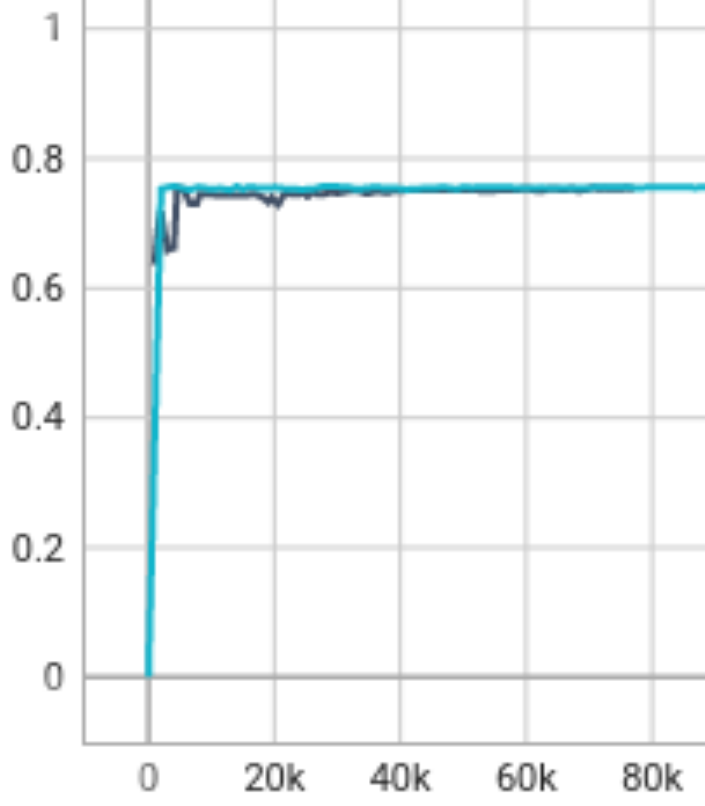}
 \caption{Comparing RNN-T (light blue) with Seq-to-Seq (dark blue) using {\em eow} encoding and hard attention. Models perform equally with similar convergence.}
 \vspace{-4mm}
 \label{fig:seqseq}
\end{figure}

We compare RNN-T and the Seq-to-Seq model using {\em eow} encoding and hard attention in Figure~\ref{fig:seqseq}. We use the same experimental setup as in the other experiments and the results in Figure ~\ref{fig:seqseq} show that both models converge similiar and obtain same performance in F1 scores.

\section{Conclusions}
We demonstrated that RNN-T works well for nested and overlapping named entity recognition and showed that the approach scales for large ontologies. The model is simple and elegant. Compared to Hierarchical CRF or reading comprehension models, only one model and one loss function is needed, making optimization much easier.

We introduced a fixed alignment loss that better utilizes human annotations and improves F1-score from $0.70$ to $0.74$. We genearlized this notion to a constrained alignment loss where users can vary the degree of reliance on the position information from the training data. This is useful in scenarios where the training data is less carefully labeled or ontologies are ambiguous.

We highlight the importance of reporting experiments on long sequences and demonstrate the sensitivity of the typical RNN-T models to long sequences. Fixed alignment loss makes the learning process much easier and allows the model to generalize to long sequences with higher accuracy.

We demonstrated that the typical RNN-T model with unconstrained loss can advantage of unlabeled data. Additionally, fixed alignment loss improves performance further from $0.74$ to $0.76$, even when the model utilizes a large pre-trained encoder.

We showed that the seq-to-seq model with explicit {\em eow} encoding plus hard attention is a special case of RNN-T models. 

Lastly, it also worth pointing out that the use of RNN-T for NER makes also deployments for spoken NLU systems easier, as the same RNN-T architecture can be used for both ASR and NER.


\newpage
\bibliographystyle{acl_natbib}
\bibliography{paper}

\begin{thebibliography}{19}
\expandafter\ifx\csname natexlab\endcsname\relax\def\natexlab#1{#1}\fi

\bibitem[{Bagby and Rao(2018)}]{Bagby2018}
Tom Bagby and Kanishka Rao. 2018.
\newblock Efficient implementation of recurrent neural network transducer in
  tensorflow.
\newblock In \emph{Proceedings of the 2018 IEEE Spoken Language Technology
  Workshop}. IEEE.

\bibitem[{Dai et~al.(2019)Dai, Yang, Yang, Carbonell, Le, and
  Salakhutdinov}]{dai2019transformerxl}
Zihang Dai, Zhilin Yang, Yiming Yang, Jaime Carbonell, Quoc Le, and Ruslan
  Salakhutdinov. 2019.
\newblock \href {https://doi.org/10.18653/v1/P19-1285} {Transformer-{XL}:
  Attentive language models beyond a fixed-length context}.
\newblock In \emph{Proceedings of the 57th Annual Meeting of the Association
  for Computational Linguistics}, pages 2978--2988, Florence, Italy.
  Association for Computational Linguistics.

\bibitem[{Devlin et~al.(2019)Devlin, Chang, Lee, and
  Toutanova}]{devlin2019bert}
Jacob Devlin, Ming-Wei Chang, Kenton Lee, and Kristina Toutanova. 2019.
\newblock \href {https://doi.org/10.18653/v1/N19-1423} {{BERT}: Pre-training of
  deep bidirectional transformers for language understanding}.
\newblock In \emph{Proceedings of the 2019 Conference of the North {A}merican
  Chapter of the Association for Computational Linguistics: Human Language
  Technologies, Volume 1 (Long and Short Papers)}, pages 4171--4186,
  Minneapolis, Minnesota. Association for Computational Linguistics.

\bibitem[{Du et~al.(2019{\natexlab{a}})Du, Chen, Kannan, Tran, Chen, and
  Shafran}]{Nan-SAT}
Nan Du, Kai Chen, Anjuli Kannan, Linh Tran, Yuhui Chen, and Izhak Shafran.
  2019{\natexlab{a}}.
\newblock \href {https://doi.org/10.18653/v1/P19-1087} {Extracting symptoms and
  their status from clinical conversations}.
\newblock In \emph{Proceedings of the 57th Annual Meeting of the Association
  for Computational Linguistics}, pages 915--925, Florence, Italy. Association
  for Computational Linguistics.

\bibitem[{Du et~al.(2019{\natexlab{b}})Du, Wang, Tran, Lee, and
  Shafran}]{du-etal-2019-learning}
Nan Du, Mingqiu Wang, Linh Tran, Gang Lee, and Izhak Shafran.
  2019{\natexlab{b}}.
\newblock \href {https://doi.org/10.18653/v1/D19-1503} {Learning to infer
  entities, properties and their relations from clinical conversations}.
\newblock In \emph{Proceedings of the 2019 Conference on Empirical Methods in
  Natural Language Processing and the 9th International Joint Conference on
  Natural Language Processing (EMNLP-IJCNLP)}, pages 4979--4990, Hong Kong,
  China. Association for Computational Linguistics.

\bibitem[{Graves(2012)}]{Graves2012}
Alex Graves. 2012.
\newblock Sequence transduction with recurrent neural networks.
\newblock \emph{CoRR}.

\bibitem[{Graves et~al.(2013)Graves, Mohamed, and Hinton}]{Graves2013}
Alex Graves, Abdel{-}rahman Mohamed, and Geoffrey Hinton. 2013.
\newblock Speech recognition with deep recurrent neural networks.
\newblock In \emph{Proceeedings of the 2013 IEEE International Conference on
  Acoustics, Speech and Signal Processing}, pages 6645--6649. {IEEE}.

\bibitem[{Huang et~al.(2011)Huang, Han, Wu, and Ioffe}]{huang-crf}
Qixing Huang, Mei Han, Bo~Wu, and Sergey Ioffe. 2011.
\newblock \href {https://doi.org/10.1109/CVPR.2011.5995571} {A hierarchical
  conditional random field model for labeling and segmenting images of street
  scenes}.
\newblock In \emph{Proceedings of IEEE Conference on Computer Vision and
  Pattern Recognition}, pages 1953--1960, Colorado Springs, USA.

\bibitem[{Kannan et~al.(2018)Kannan, Chen, Jaunzeikare, and
  Rajkomar}]{Kannan2018}
Anjuli Kannan, Kai Chen, Diana Jaunzeikare, and Alvin Rajkomar. 2018.
\newblock \href {https://doi.org/10.21437/Interspeech.2018-1318}
  {Semi-supervised learning for information extraction from dialogue}.
\newblock In \emph{Proceedings of Interspeech}, pages 2077--2081.

\bibitem[{Li et~al.(2020)Li, Feng, Meng, Han, Wu, and
  Li}]{li-etal-2020-unified}
Xiaoya Li, Jingrong Feng, Yuxian Meng, Qinghong Han, Fei Wu, and Jiwei Li.
  2020.
\newblock \href {https://doi.org/10.18653/v1/2020.acl-main.519} {A unified
  {MRC} framework for named entity recognition}.
\newblock In \emph{Proceedings of the 58th Annual Meeting of the Association
  for Computational Linguistics}, pages 5849--5859, Online. Association for
  Computational Linguistics.

\bibitem[{Manning and Sch{\"u}tze(1999)}]{manning99foundations}
Christopher~D. Manning and Hinrich Sch{\"u}tze. 1999.
\newblock \href {http://nlp.stanford.edu/fsnlp/} {\emph{Foundations of
  Statistical Natural Language Processing}}.
\newblock The {MIT} Press, Cambridge, Massachusetts.

\bibitem[{McCann et~al.(2018)McCann, Keskar, Xiong, and
  Socher}]{mccann2018natural}
Bryan McCann, Nitish~Shirish Keskar, Caiming Xiong, and Richard Socher. 2018.
\newblock \href {http://arxiv.org/abs/1806.08730} {The natural language
  decathlon: Multitask learning as question answering}.

\bibitem[{Shafran et~al.(2020)Shafran, Du, Tran, Perry, Keyes, Knichel, Domin,
  Huang, Chen, Li, Wang, El~Shafey, Soltau, and
  Paul}]{shafran-etal-2020-medical}
Izhak Shafran, Nan Du, Linh Tran, Amanda Perry, Lauren Keyes, Mark Knichel,
  Ashley Domin, Lei Huang, Yu-hui Chen, Gang Li, Mingqiu Wang, Laurent
  El~Shafey, Hagen Soltau, and Justin~Stuart Paul. 2020.
\newblock \href {https://www.aclweb.org/anthology/2020.lrec-1.250} {The medical
  scribe: Corpus development and model performance analyses}.
\newblock In \emph{Proceedings of the 12th Language Resources and Evaluation
  Conference}, pages 2036--2044, Marseille, France. European Language Resources
  Association.

\bibitem[{Sim et~al.(2017)Sim, Narayanan, Bagby, Sainath, and
  Bacchiani}]{Sim2017}
Khe~Chai Sim, Arun Narayanan, Tom Bagby, Tara~N Sainath, and Michiel Bacchiani.
  2017.
\newblock Improving the efficiency of forward-backward algorithm using batched
  computation in tensorflow.
\newblock In \emph{2017 IEEE Automatic Speech Recognition and Understanding
  Workshop}, pages 258--264. IEEE.

\bibitem[{Soltau et~al.(2021)Soltau, Wang, Shafran, and Shafey}]{Soltau2021}
Hagen Soltau, Mingqiu Wang, Izhak Shafran, and Laurent~El Shafey. 2021.
\newblock \href {https://doi.org/10.21437/Interspeech.2021-691} {{Understanding
  Medical Conversations: Rich Transcription, Confidence Scores \& Information
  Extraction}}.
\newblock In \emph{Proceedings of Interspeech}, pages 4418--4422.

\bibitem[{Strakov{\'a} et~al.(2019)Strakov{\'a}, Straka, and
  Hajic}]{strakova-etal-2019-neural}
Jana Strakov{\'a}, Milan Straka, and Jan Hajic. 2019.
\newblock \href {https://doi.org/10.18653/v1/P19-1527} {Neural architectures
  for nested {NER} through linearization}.
\newblock In \emph{Proceedings of the 57th Annual Meeting of the Association
  for Computational Linguistics}, pages 5326--5331, Florence, Italy.
  Association for Computational Linguistics.

\bibitem[{Vaswani et~al.(2017)Vaswani, Shazeer, Parmar, Uszkoreit, Jones,
  Gomez, Kaiser, and Polosukhin}]{Vaswani2017}
Ashish Vaswani, Noam Shazeer, Niki Parmar, Jakob Uszkoreit, Llion Jones,
  Aidan~N Gomez, {\L}ukasz Kaiser, and Illia Polosukhin. 2017.
\newblock Attention is all you need.
\newblock In \emph{Advances in neural information processing systems},
  volume~30, pages 5998--6008.

\bibitem[{Virpioja et~al.(2013)Virpioja, Smit, Gr{\"o}nroos, and
  Kurimo}]{Virpioja2013}
Sami Virpioja, Peter Smit, Stig-Arne Gr{\"o}nroos, and Mikko Kurimo. 2013.
\newblock \href {http://urn.fi/URN:ISBN:978-952-60-5501-5} {Morfessor 2.0:
  Python implementation and extensions for morfessor baseline}.
\newblock Technical report, Aalto University.

\bibitem[{Zhang et~al.(2020)Zhang, Lu, Sak, Tripathi, McDermott, Koo, and
  Kumar}]{qian2020}
Qian Zhang, Han Lu, Hasim Sak, Anshuman Tripathi, Erik McDermott, Stephen Koo,
  and Shankar Kumar. 2020.
\newblock Transformer transducer: {A} streamable speech recognition model with
  transformer encoders and {RNN-T} loss.
\newblock In \emph{Proceedings of the 2020 IEEE International Conference on
  Acoustics, Speech and Signal Processing}, pages 7829--7833. IEEE.

\end{thebibliography}

\end{document}